\documentclass[12pt]{article}

\usepackage{mathptmx}       
\usepackage{helvet}         
\usepackage{courier}        
\usepackage{makeidx}         
\usepackage{graphicx}        
\usepackage{multicol}        
\usepackage[bottom]{footmisc}

\usepackage{natbib}
\usepackage{graphicx} 
\usepackage[utf8]{inputenc}
\usepackage{float}
\usepackage{refcount}
\usepackage{amsfonts}

\def\E{\mathbb{E}}

\begin{document}
\title{Evolving Culture vs Local Minima}
\author{Yoshua Bengio\\
Department of computer science and operations research, U. Montreal}

\date{}

\newcounter{O}
\setcounter{O}{0}
\newcounter{H}
\setcounter{H}{0}
\def\nO#1{\refstepcounter{O}\label{#1}O\ref{#1}}

\maketitle

\abstract{
  We propose a theory that relates difficulty of learning in deep
  architectures to culture and language. It is articulated around
  the following hypotheses: (1) learning in an individual 
  human brain is hampered by the presence of effective local minima; 
  (2) this optimization difficulty is particularly important
  when it comes to learning higher-level abstractions, i.e., concepts
  that cover a vast and highly-nonlinear span of sensory configurations;
  (3) such high-level abstractions are best represented in
  brains by the composition of many levels of representation, i.e.,
  by deep architectures;
  (4) a human brain can learn such high-level abstractions
  if guided by the signals produced by other humans, which act as hints
  or indirect supervision for these high-level abstractions;
  and (5), language and the recombination
  and optimization of mental concepts provide an efficient evolutionary
  recombination operator, and this  
  gives rise to rapid search in the space of communicable ideas
  that help humans build up better 
  high-level internal representations of their world.
  These hypotheses put together imply
  that human culture and the evolution of ideas
  have been crucial to counter an optimization difficulty:
  this optimization difficulty 
  would otherwise make it very difficult for human brains
  to capture high-level knowledge of the world.
  The theory is grounded in experimental observations of the 
  difficulties of training deep artificial neural networks. Plausible
  consequences of this theory for the efficiency of cultural evolution
  are sketched.
}

\section{Introduction}

Interesting connections can sometimes be made at the interface between
artificial intelligence research and the sciences that aim to understand
human brains, cognition, language, or society.  The aim of this paper is
to propose and elaborate a theory at this interface, inspired by
observations rooted in machine learning research, on so-called Deep
Learning\footnote{See~\citet{Bengio-2009} for a review of Deep Learning
  research, which had a breakthrough in 2006~\citep{Hinton06,Bengio-nips-2006,ranzato-07}}. 
Deep Learning techniques aim at training models with many
levels of representation, a hierarchy of features and concepts, 
such as can be implemented with artificial neural networks with many
layers. A {\em deep architecture} has typically more than 2 or 3 
trained levels of representation, and in fact we consider that
a {\em deep learning algorithm} can {\em discover} the appropriate
number of levels of representation, based on the training data.
The visual cortex is believed to have between 5 and 10 such levels.
Theoretical arguments
have also been made to suggest that deep architectures are necessary to
efficiently represent the kind of high-level concepts required for
artificial intelligence~\citep{Bengio+chapter2007}.
This paper starts from
experimental observations of the difficulties in training deep
architectures~\citep{Erhan+al-2010}, 
and builds a theory of the role of cultural evolution to
reduce the difficulty of learning high-level abstractions.
The gist of this theory is that training deep architectures
such as those found in the brain is difficult because of
an optimization difficulty (local minima), but that the cultural
evolution of ideas can serve as a way for a whole population
of humans, over many generations, to efficiently discover
better solutions to this optimization problem.

\section{Neural Networks and Local Minima}

\subsection{Neural Networks}

Artificial neural networks are computational architectures and learning
algorithms that are inspired from the computations believed to take place
in the biological neural networks of the brain~\citep{Arbib-1995}.
The dominant and most successful approaches to training artificial
neural networks are all based on the idea that {\em learning can proceed
by gradually optimizing a criterion}~\citep{Rumelhart86}. A neural network typically
has free parameters, such as the synaptic strengths associated with
connections between neurons. Learning algorithms formalize the
computational mechanism for changing these parameters so as to
take into account the evidence provided by observed (training) examples.
Different learning algorithms
for neural networks differ in the specifics of the criterion and
how they optimize it, often approximately because no analytic 
and exact solution is possible.
On-line learning, which is most plausible for biological
organisms, involves changes in the parameters either after each
example has been seen or after a small batch of examples has
been seen (maybe corresponding to a day's worth of experience).

\subsection{Training Criterion}

In the case of biological organisms, one could imagine that
the ultimate criterion involves the sum of expected future
rewards (survival, reproduction, and other innately defined 
reward signals such as hunger, thirst, and the need to sleep)\footnote{Note
that the rewards received by an agent depend on the tasks that it faces,
which may be different depending on the biological and social niche that
it occupies.}. 
However, intermediate criteria typically involve modeling
the observations from the senses, i.e., improving the prediction
that could be made of any part of the observed sensory input given any other part,
and improving the prediction of future observations given the
past observations. Mathematically, this can often be captured
by the statistical criterion of maximizing likelihood, i.e.,
of maximizing the probability that the model implicitly or
explicitly assigns to new observations.

\subsection{Learning}

Learners can exploit observations (e.g., from their sensors of
the real world) in order to construct functions that capture
some of the statistical relationships between the observed variables.
For example, learners can build predictors of future events given
past observations, or associate what is observed through different
modalities and sensors. This may be used by the learner to predict
any unobserved variable given the observed ones.
The learning problem can be formalized as follows. Let $\theta$
be a vector of parameters that are free to change while learning
(such as the synaptic strengths of neurons in the brain).
Let $z$ represent an example, i.e., a measurement of the variables
in the environment which are relevant to the learning agent.
The agent has seen a past history $z_1,z_2,\ldots,z_t$,
which in realistic cases also depends on the actions of the agent.
Let $E(\theta,z)$ be a measurement of an error or loss to be minimized,
whose future expected value is the criterion to be minimized.
In the simple case\footnote{stationary i.i.d case where examples independently
come from the same stationary distribution $P$}
where we ignore the effect of current actions
on future rewards but only consider the value of a particular
solution to the learning problem over the long term, the 
objective of the learner is to minimize the criterion
\begin{equation}
\label{eq:C}
  C(\theta) = \int P(z) E(\theta,z) dz = \E[E(\theta,Z)]
\end{equation}
which is the expected future error, with
$P(z)$ the unknown probability distribution from
which the world generates examples for the learner. In the
more realistic setting of reinforcement learning~\citep{Sutton+Barto-98},
the objective of the learner is often formalized as the maximization
of the expected value of the weighted sum of future rewards,
with weights that decay as we go further into the future 
(food now is valued more than food tomorrow, in general).
Note that the training criterion we define here is called
{\em generalization error} because it is the expected error on new
examples, not the error measured on past training examples
(sometimes called training error). Under the stationary
i.i.d. hypothesis, the expected future
reward can be estimated by the ongoing online error, which
is the average of rewards obtained by an agent. In any case,
although the training criterion cannot be computed
exactly (because $P(\cdot)$ is unknown to the learner),
the criterion $C(\cdot)$ can be approximately 
minimized\footnote{In many machine learning algorithms,
one minimizes the training error plus a regularization penalty
which prevents the learner from simply learning the training examples by heart
without good generalization on new examples.}
by stochastic gradient descent (as well as other gradient-based
optimization techniques): the learner just needs to 
estimate the gradient $\frac{\partial E(\theta,z)}{\partial \theta}$
of the example-wise error $E$ with respect to the parameters, i.e., estimate the
effect of a change of the parameters on the immediate error. Let $g$
be such an estimator (e.g., if it is unbiased 
then $\E[g]=\E[\frac{\partial E(\theta,z)}{\partial \theta}]$).
For example, $g$ could be based on a single example or a day's
worth of examples.

Stochastic gradient descent proceeds by small steps of
the form
\begin{equation}
\label{eq:sgd}
  \theta \leftarrow \theta - \alpha g
\end{equation}
where $\alpha$ is a small constant called learning rate or gain.
Note that if new examples $z$ are continuously sampled from the unknown
distribution $P(z)$, the instantaneous online gradient $g$ is an unbiased
estimator of the generalization error gradient (which is the integral
of $g$ over $P$), i.e., an online learner is directly optimizing
generalization error.

Applying these ideas to the context of
biological learners gives the hypothesis that follows.

%
%
%
%
%
%
%
%
%

\subsection{What do brains optimize?}

\begin{center}
\framebox{
\begin{minipage}{0.8\linewidth}
{\bf Optimization Hypothesis.} When the brain of a single biological
agent learns, it performs an approximate optimization with respect to
some endogenous objective.
\end{minipage}
}\\
\end{center}

Here note that we refer to a {\em single} learning agent because
we exclude the effect of interactions between learning agents, 
like those that occur because of communication between humans in
a human society. Later we will advocate that in fact when one takes
into account the learning going on throughout a society, the optimization
is not just a local descent but involves a global parallel search similar
to that performed by evolution and sexual reproduction. 

Note that the criterion we have in mind here is not specialized to a single
task, as is often the case in applications of machine learning. Instead, a
biological learning agent must make good predictions in all the contexts
that it encounters, and especially those that are more relevant to its
survival. Each type of context in which the agent must take a
decision corresponds to a ``task''. The
agent needs to ``solve'' many tasks, i.e. perform {\em multi-task
  learning}, {\em transfer learning} or {\em self-taught 
learning}~\citep{caruana93a,RainaR2007}. All the tasks faced by the learner
share the same underlying
``world'' that surrounds the agent, and brains probably take advantage of
these commonalities. This may explain how brains can sometime learn a new
task from a handful or even just one example, 
something that seems almost impossible with standard single-task 
learning algorithms.

Note also that biological agents probably need to address multiple
objectives together. However, in practice, since the same brain must take
the decisions that can affect all of these criteria, these cannot be
decoupled but they can be lumped into a single criterion with appropriate
weightings (which may be innate and chosen by evolution). For example, it
is very likely that biological learners must cater both to a ``predictive''
type of criterion (similar to the data-likelihood used in statistical
models or in {\em unsupervised learning} algorithms) and a ``reward'' type of
criterion (similar to the rewards used in reinforcement learning
algorithms). The former explains curiosity and our ability to make sense
of observations and learn from them even when we derive no immediate or foreseeable
benefit or loss. The latter is clearly crucial for survival, as biological
brains need to focus their modeling efforts on what matters most to
survival. Unsupervised learning is a way for a learning agent to
prepare itself for any possible task in the future, by extracting
as much information as possible from what it observes, i.e., figuring
out the unknown explanations for what it observes.

One issue with an objective defined in terms of an animal survival and
reproduction (presumably the kinds of objectives that make sense in
evolution) is that it is not well defined: it depends on the behaviors of
other animals and the whole ecology and niche occupied by the animal of
interest. As these change due to the ``improvements'' made by other animals
or species through evolution or learning, the individual animal or
species's objective of survival also changes. This feedback loop means
there isn't really a static objective, but a complicated dynamical system,
and the discussions regarding the complications this brings are beyond the
scope of this paper. However, it is interesting to note that there is one
component of an animal's objective (and certainly of many humans'
objective, especially scientists) that is much more stable: it is the
``unsupervised learning'' objective of ``understanding how the world
ticks''.

\subsection{Local Minima}

Stochastic gradient descent is one of many 
optimization techniques that perform a {\em local descent}:
starting from a particular configuration of the parameters
(e.g. a configuration of the brain's synapses), one makes small
gradual adjustments which in average tend to improve the
expected error, our training criterion. The theory proposed
here relies on the following hypothesis:\\

\begin{center}
\framebox{
\begin{minipage}{0.8\linewidth}
{\bf Local Descent Hypothesis.} When the brain of a single biological
agent learns, it relies
on approximate local descent in order to
gradually improve itself.
\end{minipage}
}\\
\end{center}

The main argument in favor of this hypothesis would be based on the
assumption that although our state of mind (firing pattern of neurons)
changes quickly, synaptic strengths and neuronal connectivity only change
gradually.

If the learning algorithm is a form of
stochastic gradient descent (as eq.~\ref{eq:sgd} above), 
where $g$ approximates the gradient (it may even have a bias),
and if $\alpha$
is chosen small enough (compared to the largest second derivatives
of $C$), then $C$ will gradually decrease with high probability, and
if $\alpha$ is gradually decreased at an appropriate rate
(such as $1/t$), then the learner will converge towards a 
{\em local minimum} of $C$. The proofs are usually for the
unbiased case~\citep{bottou-mlss-2004}, 
but a small bias is not necessarily very
hurtful, as shown for Contrastive Divergence~\citep{Perpinan+Hinton-2005,Yuille2005},
especially if the magnitude of the bias also decreases
as the gradient decreases (stochastic approximation
convergence theorem~\citep{Yuille2005}). 

Note that in this paper we are talking about {\em local minima of
  generalization error}, i.e., with respect to expected future rewards. In
machine learning, the terms ``optimization'' and ``local minimum'' are
usually employed with respect to a training criterion formed by the error
on training examples (training error), which are those seen in the past by
the learner, and on which it could possibly {\em overfit} (i.e. perform
apparently well even though generalization error is poor).

\begin{figure}[h]
\hspace*{-6mm} %
\resizebox{0.4\textwidth}{!}{\includegraphics{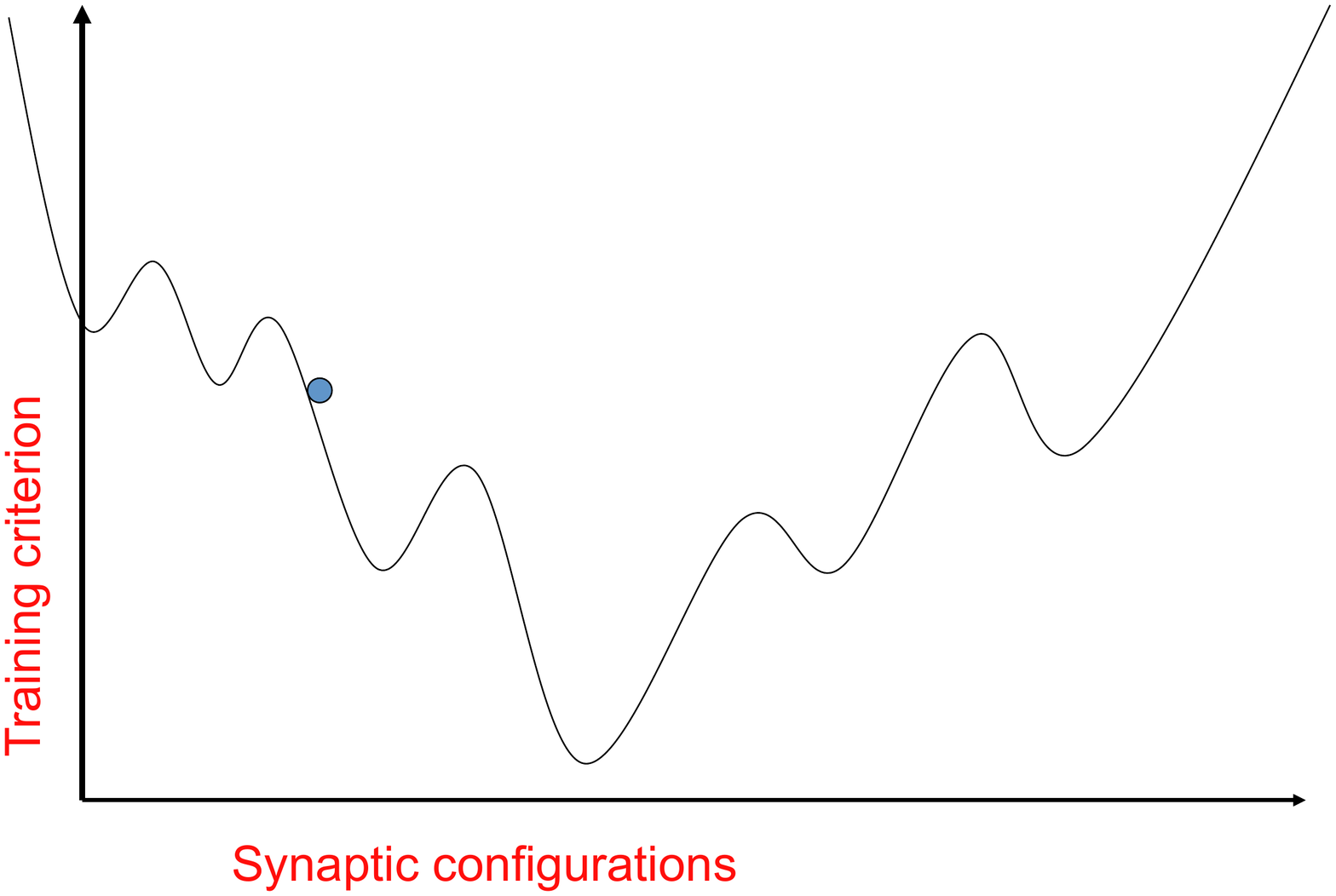}} %
\hspace*{-16.5mm} %
\resizebox{0.4\textwidth}{!}{\includegraphics{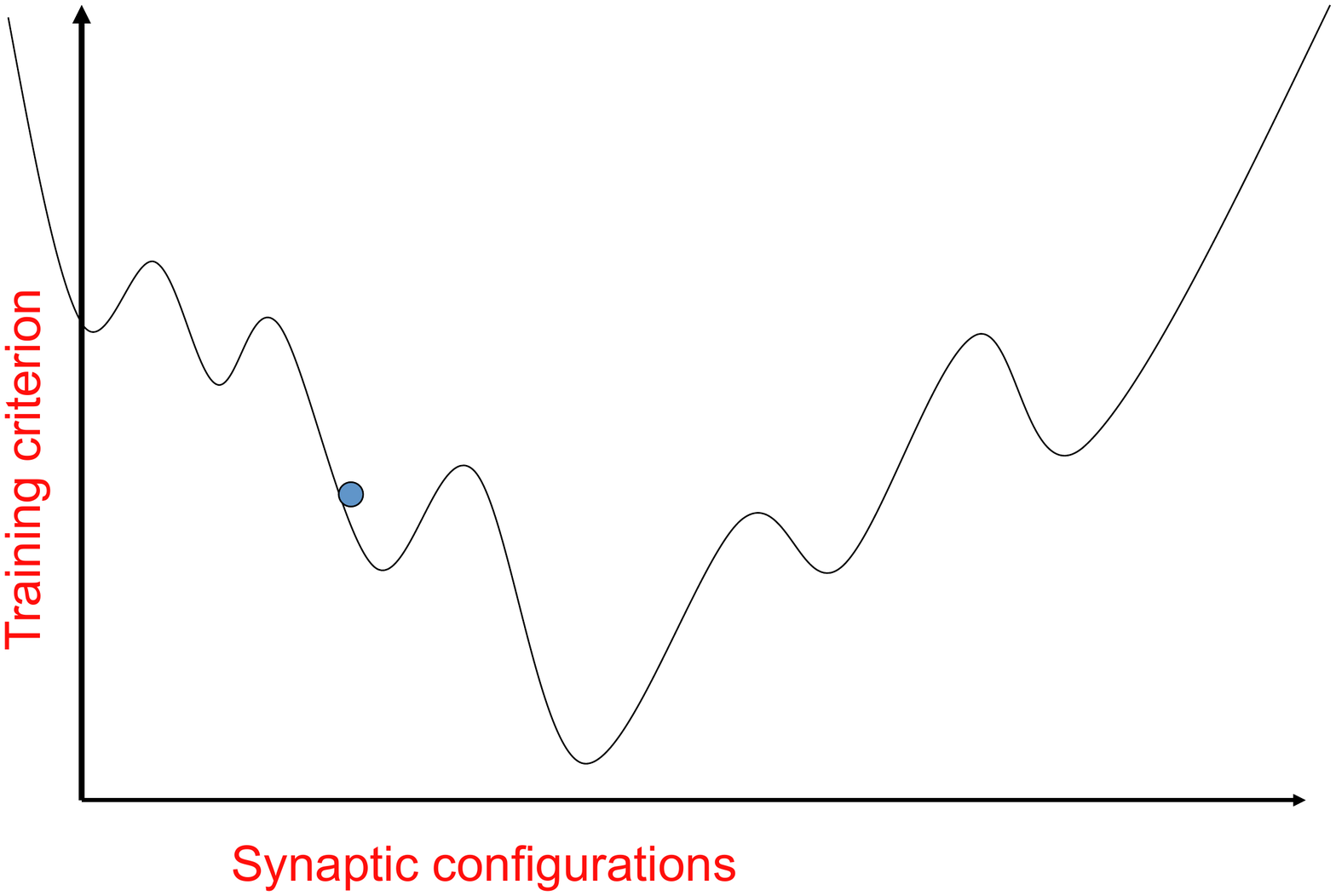}} %
\hspace*{-16.5mm} %
\resizebox{0.4\textwidth}{!}{\includegraphics{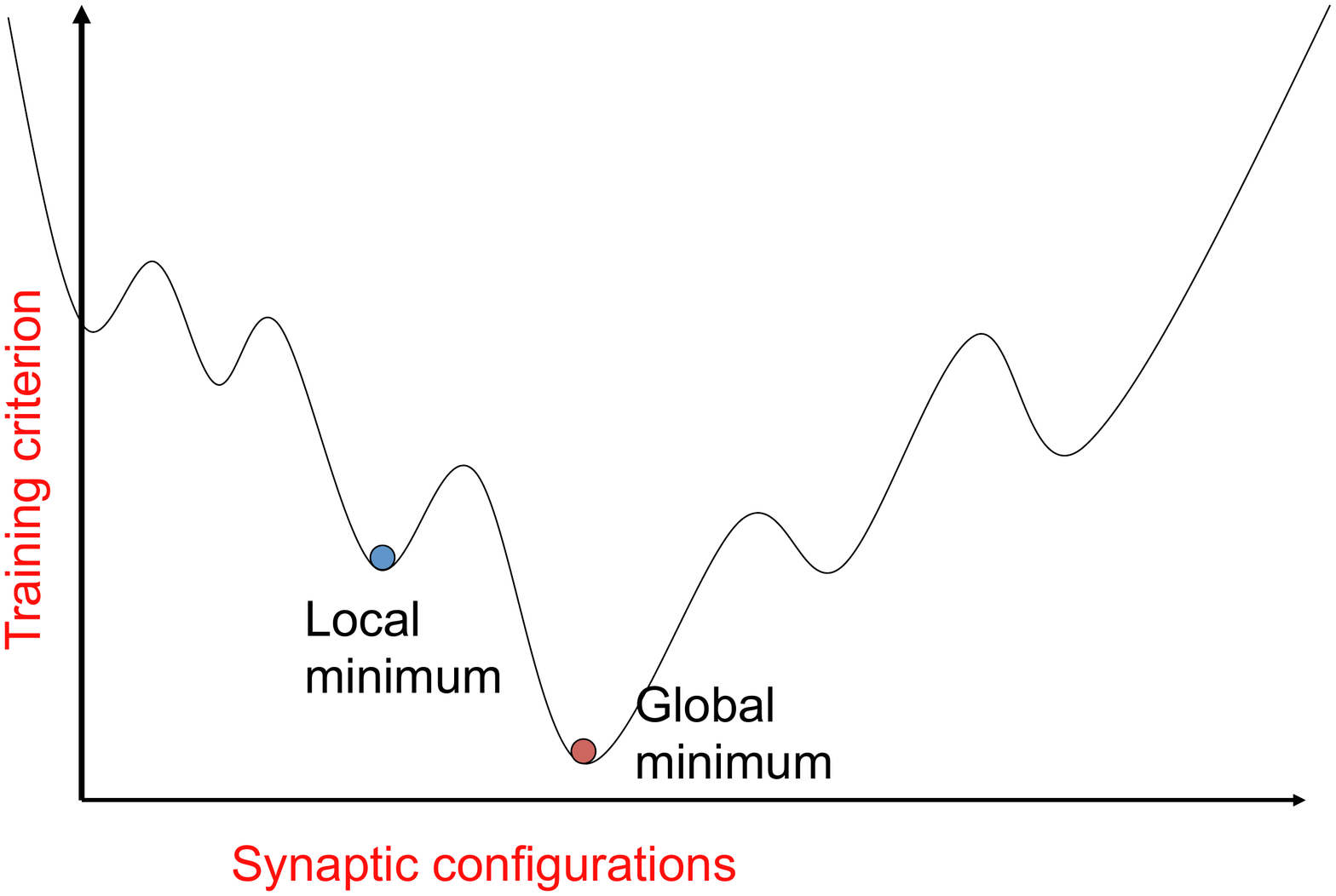}} %
\hspace*{-16.5mm} %
\caption{\sl Illustration of learning that proceeds by local descent, and
can get stuck near a local minimum (going from left figure to right figure). 
The horizontal axis represents
the space of synaptic configurations (parameters of the learner), while
the vertical axis represents the training criterion (expected future error).
The ball represents the learner's current state, which tends to go
downwards (improving the expected error).
Note that the space of synaptic configurations is huge (number of synaptic
connections on the order of 100 trillion in humans) but represented here
schematically with a single dimension, the horizontal axis.
\label{fig:localmin}
}
\end{figure}

\subsection{Effective Local Minima}

As illustrated in
Figure~\ref{fig:localmin}, {\em a local minimum is a configuration
of the parameters such that no small change can yield an
improvement of the training criterion.}
A consequence of the {\bf Local Descent Hypothesis}, if it is true, is therefore that
biological brains would be likely to stop improving after some point,
after they have sufficiently approached a local minimum. In practice,
if the learner relies on a stochastic gradient estimator (which
is the only plausible hypothesis we can see, because no biological learner has access to
the full knowledge of the world required to directly estimate $C$),
it will continue to change due to the stochastic nature of the
gradient estimator (the training signal), hovering stochastically
around a minimum. It is also quite possible that biological learners
do not have enough of a lifetime to really get very close to an actual
local minimum, but what is plausible is that they get to a point where
progress is very slow (so slow as to be indistinguishable from random
hovering near a minimum). In practice, when one trains an artificial
neural network with a learning algorithm based on stochastic gradient
descent, one often observes that training {\em saturates}, i.e., no
more observable progress is seen in spite of the additional examples
being shown continuously. The learner appears stuck near a local
minimum. Because it is difficult to verify that a learner is really
near a local minimum, we call these {\em effective local minima}.
We call it effective because it is due to the limitations of the
optimization procedure (e.g., stochastic gradient descent) and not
just to the shape of the training criterion as a function of the parameters.
The learner equiped with its optimization procedure is stuck
in an effective local minima and it looks like it is stuck in an actual local minima
(it may also be an actual local minima).
It may happen that the training criterion is a complicated 
function of the parameters, such that stochastic gradient descent is
sometimes practically stuck in a place in which it is not possible
to improve in {\em most directions}, but from where other more powerful descent
methods could escape~\citep{martens2010hessian}. 

\subsection{Inference}

Many learning algorithms involve {\em latent variables}, which can be
understood as associated with particulars factors that can contribute
to ``explain'' each other and ``explain'' the current and recent
observations. These latent variables are encoded in the activation
of {\em hidden units} (neurons that are neither inputs nor outputs).
One can think of a particular configuration of these latent or hidden
variables as corresponding to a {\em state of mind}.
In a Boltzmann machine~\citep{Hinton-Boltzmann,Ackley85,Salakhutdinov2009},
when an input is presented, many configurations of these latent variables 
are possible and an {\em inference mechanism} normally takes place
in order to explore possible configurations of the latent variables
which ``fit well'' with the observed input. This inference is often
iterative, although it can be approximated or initialized in a single bottom-up
pass~\citep{Salakhutdinov+Larochelle-2010} from perception to state-of-mind. 
Inference can be
stochastic (neurons randomly choose their state with a probability
that depends on the state of the others, and such that more
probable configurations of neuron activations are sampled accordingly
more often) or deterministic (through an iterative process that
can sometimes correspond to an optimization, gradually changing
the configuration of neurons towards one that agrees more with
the observed input percept). Whereas learning in brains (besides the simple memorization
of facts and observed events) occurs on a scale of minutes,
hours or days, inference (changes in the state of mind) occurs on the scale of a fraction of
a second or few seconds. Whereas learning is
probably gradual, stochastic inference can quickly jump from
one thought pattern to another in less than a second. In models
such as the Boltzmann machine, learning requires inference as
an inner loop: patterns of latent variables (hidden units,
high-level concepts) that fit well with the observed data
are reinforced by the changes in synaptic weights that follow.
One should not confuse local minima in synaptic weights
with local minima (or the appearance of being stuck) in inference.
Randomness or association with new stimuli can change the state
of our inference for past inputs and give us the impression
that we are not stuck anymore, that we have escaped a local
minimum, but that regards the inference process, not necessarily
the learning process (although it can certainly help it).

\section{High-Level Abstractions and Deep Architectures}

Deep architectures~\citep{Bengio-2009} are parametrized families of
functions which can be used to model data using multiple levels of
representation.  In deep neural networks, each level is associated with a
group of neurons (which in the brain could correspond to an area, such as
areas V1, V2 or IT of the visual cortex). During sensory perception in
animal brains, information travels quickly from lower (sensory) levels to
higher (more abstract) levels, but there are also many feedback connections
(going from higher to lower levels) as well as lateral connections (between
neurons at the same level).  Each neuron or group of neurons can be thought
of as capturing a concept or feature or aspect, and being activated when
that concept or feature or aspect is present in the sensory input, or when
the model is generating an internal configuration (a ``thought'' or
``mental image'') that includes that concept or feature or aspect. Note
that very few of these features actually come to our consciousness, because
most of the inner workings of our brains are not directly accessible (or
rarely so) to our consciousness. Note also that a particular linguistic
concept may be represented by many neurons or groups of neurons, activating
in a particular pattern, and over different levels (in fact so many neurons
are activated that we can see whole regions being activated with brain
imaging, even when a single linguistic concept is presented as
stimulus). These ideas were introduced as central to connectionist 
approaches~\citep{Rumelhart86,Hinton86b,Hinton89b} to
cognitive science and artificial neural networks, with the concept of
{\em distributed representation}: what
would in most symbolic systems be represented by a single ``on/off'' bit
(e.g., the symbol for 'table' is activated) is associated in the brain with
a large number of neurons and groups of neurons being activated together in
a particular pattern. In this way, concepts that are close semantically,
i.e., share some attributes (e.g. represented by a group of neurons), can
have an overlap in their brain representation, i.e., their corresponding
patterns of activation have ``on'' bits in many of the same places.

\subsection{Efficiency of Representation}
\label{sec:efficiency-repr}

Deeper architectures can be much more efficient in terms of representation
of functions (or distributions) than shallow ones, as shown with theoretical
results where for specific families of functions a too shallow architecture
can require exponentially more resources than 
necessary~\citep{Yao85,Hastad86,Hastad91,Bengio+chapter2007,Bengio-2009,Bengio-decision-trees10,Bengio+Delalleau-ALT-2011}.
The basic intuition why this can be true is that in a deep architecture
there is {\em re-use} of parameters and {\em sharing} of sub-functions to
build functions.
We do not write computer programs with a single main program:
instead we write many subroutines (functions) that can call other subroutines, and this
nested re-use provides not only flexibility but also great expressive power.
However, this greater expressive power may come at the price of making the learning
task a more difficult optimization problem. Because the lower-level features can be used
in many ways to define higher-level features, the interactions between parameters
at all levels makes the optimization landscape much more complicated. 
At the other extreme, many shallow methods are associated with a convex
optimization problem, i.e., with a single minimum of the training criterion.

\begin{figure}[H]
\vspace*{-1mm}
\centerline{
\resizebox{0.7\textwidth}{!}{\includegraphics{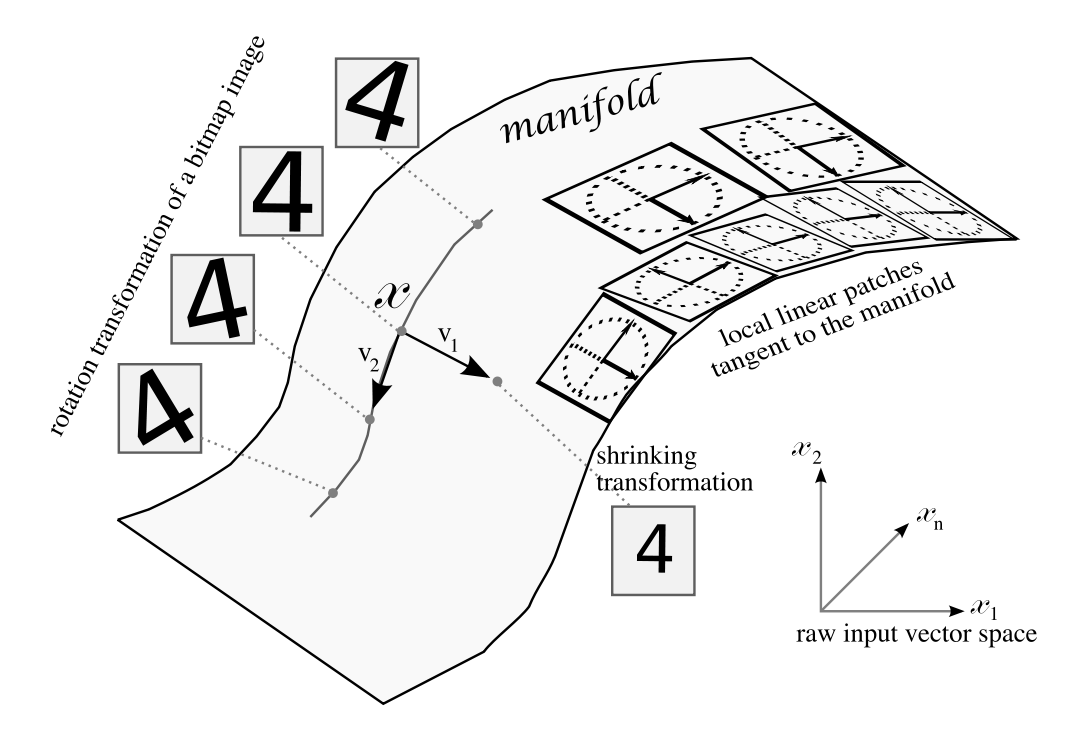}} %
}
\caption{\sl Example of a simple manifold in the space of images,
associated with a rather low-level concrete
concept, corresponding to rotations and shrinking of a {\bf specific} instance
of the image of a drawn digit 4. Each point on the manifold corresponds to an image which
is obtained by rotating or translating or scaling
another image on the manifold. The set of
points in the manifold defines a concrete concept associated with the 
drawing of a 4 of a particular shape irrespective of its position, angle and scale. 
Even learning
such simple manifolds is difficult, but learning the much more convoluted and
higher-dimensional manifolds of more abstract concepts is much harder.
\label{fig:manifold}
}
\end{figure}

\subsection{High-Level Abstractions}

We call {\em high-level abstraction} the kind of concept
or feature that could be computed efficiently only through a deep structure in the
brain (i.e., by the sequential application of several different
transformations, each associated with an area of the brain or large group
of neurons). An edge detector in an image seen by the eye can be
computed by a single layer of neurons from raw pixels, using Gabor-like
filters. This is a very low-level abstraction. Combining several such
detectors to detect corners, straight line segments, curved line segments,
and other very local but simple shapes can be done by one or two more
layers of neurons, and these can be combined in such a way as to be
locally insensitive to small changes in position or angle. Consider a
hierarchy of gradually more complex features, constructing detectors for very abstract
concepts which are activated whenever any stimulus within 
a very large set of possible input stimuli
are presented. For a higher-level abstraction, this set of stimuli
represents a highly-convoluted set of points, a highly curved manifold.
We can picture such a manifold if we restrict ourselves to a very
concrete concept, like the image of a specific object (the digit 4,
as in Figure~\ref{fig:manifold}) on a uniform background. The only factors 
that can vary here are due to object constancy; they correspond
to changes in imaging geometry (location and orientation
of the object with respect to the eye) and lighting, and we can
use mathematics to help us make sense of such manifolds. Now
think about all the images which can elicit a thought of a more abstract
concept, such as 
``human'', or even more abstract, all the contexts which can elicit a thought
of the concept ``Riemann integral''. These contexts and images associated with
the same high-level concept can be very different from 
each other, and in many complicated ways, for which scientists do not know how
to construct the associated manifolds. Some concepts are clearly higher-level
than others, and often we find that higher-level concepts can be defined
in terms of lower-level ones, hence forming a hierarchy which is reminiscent
of the kind of hierarchy that we find current deep learning algorithms
to discover~\citep{HonglakL2009}. This discussion brings us to the
formulation of a hypothesis about high-level abstractions and their
representation in brains.\\

\begin{center}
\framebox{
\begin{minipage}{0.8\linewidth}
{\bf Deep Abstractions Hypothesis.} Higher-level abstractions in brains are represented
by deeper computations (going through more areas or more computational steps in sequence
over the same areas).
\end{minipage}
}\\
\end{center}

\section{The Difficulty of Training Deep Architectures}

There are a number of results in the machine learning literature that
suggest that training a deeper architecture is often more difficult than
training a shallow one, in the following sense. When trying to train all
the layers together with respect to a joint criterion such as the
likelihood of the inputs or the conditional likelihood of target classes
given inputs, results can be worse than when training a shallow model,
or more generally, one may suspect that current training procedures for
deep networks underuse the representation potential and the parameters available,
which may correspond to a form of underfitting\footnote{although
it is always possible to trivially overfit the top two layers of a deep network
by memorizing patterns, this may still happen with very poor training of
lower levels, corresponding to poor representation learning.}
and inability at learning
very high-level abstractions.

\subsection{Unsupervised Layer-Wise Pre-training}

The first results of that nature appear in~\citet{Bengio-nips-2006,ranzato-07}, where
the same architecture gives very different results depending on the
initialization of the network weights, either purely randomly, or based on
{\em unsupervised layer-wise pre-training}.  The idea of the layer-wise
pre-training
scheme~\citep{Hinton-Science2006,Hinton06,Bengio-nips-2006,ranzato-07} is
to train each layer with an unsupervised training criterion, so that it
learns a new representation, taking as input the representation of the
previous layer. Each layer is thus trained in sequence one after the
other. Although this is probably not biologically plausible as such, what
would be plausible is a mechanism for providing an unsupervised signal at
each layer (group of neurons) that makes it learn to better capture the
statistical dependencies in its inputs. That layer-local signal could still
be combined with a global training criterion but might help to train deep
networks if there is an optimization difficulty in coordinating the
training of all layers simultaneously. Another indication that a
layer-local signal can help to train deep networks came from the work of
~\citet{WestonJ2008}, where the unsupervised layer-local signal was
combined with a supervised global signal that was propagated through the
whole network. This observation of the advantage brought by layer-local
signals was also made in the context of purely unsupervised learning of a
deep stochastic network, the Deep Boltzmann
Machine~\citep{Salakhutdinov2009}. By pre-training each layer as a
Restricted Boltzmann Machine (RBM)\footnote{which ignores the interaction
  with the other levels, except for receiving input from the level below.}
before optimizing a Deep Boltzmann Machine (DBM) that comprises all the
levels, the authors are able to train the DBM, whereas directly training it
from random initialization was problematic. We summarize several of the
above results in the deep learning literature with the following {\bf
  Observation \nO{obs:layer-hints}}: training deep architectures is easier
if hints are provided about the function that intermediate levels should
compute~\citep{Hinton06,WestonJ2008,Salakhutdinov2009,Bengio-2009}.  This
is connected to an even more obvious {\bf Observation \nO{obs:sup-easier}},
from the work on artificial neural networks: it is much easier to teach a
network with supervised learning (where we provide it examples of when a
concept is present and when it is not present in a variety of examples)
than to expect unsupervised learning to discover the concept (which may
also happen but usually leads to poorer renditions of the concept).

\begin{figure}[H]
\vspace*{-2mm}
\centerline{
\resizebox{0.51\textwidth}{!}{\includegraphics{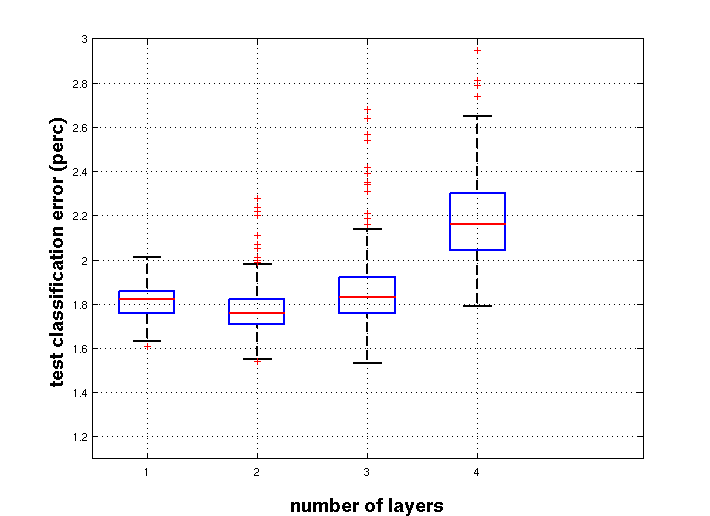}} 
\hspace*{-8mm} 
\resizebox{0.51\textwidth}{!}{\includegraphics{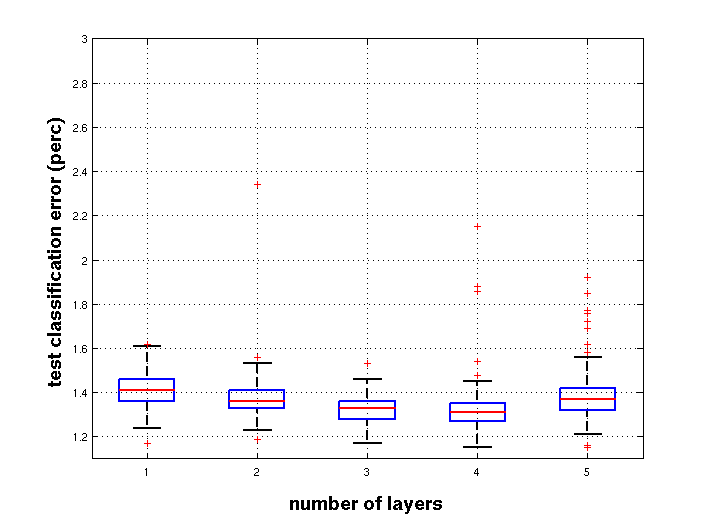}}
}
\vspace*{-2mm}
\caption{\sl Effect of depth on generalization error, {\bf without} layer-wise unsupervised
pre-training (left) and {\bf with} (right). The training problem becomes more
difficult for deeper nets, and using a layer-local cue to initialize each
level helps to push the difficulty a bit farther and improve error rates.
\label{fig:worse-deeper}
}
\end{figure}

\subsection{More Difficult for Deeper Architectures and More Abstract Concepts}

Another clue to this training difficulty came in later
studies~\citep{Larochelle-jmlr-toappear-2008,Erhan+al-2010} showing that
directly training all the layers together would not only make it difficult
to exploit all the extra modeling power of a deeper architecture but would
actually get worse results\footnote{Results got worse in terms of
  generalization error, while training error could be small thanks to
  capacity in the top few layers.} {\em as the number of layers is
  increased}, as illustrated in Figure~\ref{fig:worse-deeper}.  We call
this {\bf Observation \nO{obs:deep-harder}}.

\begin{figure}[H]
\begin{center}
\vspace*{-3mm}
\resizebox{0.6\textwidth}{!}{\includegraphics{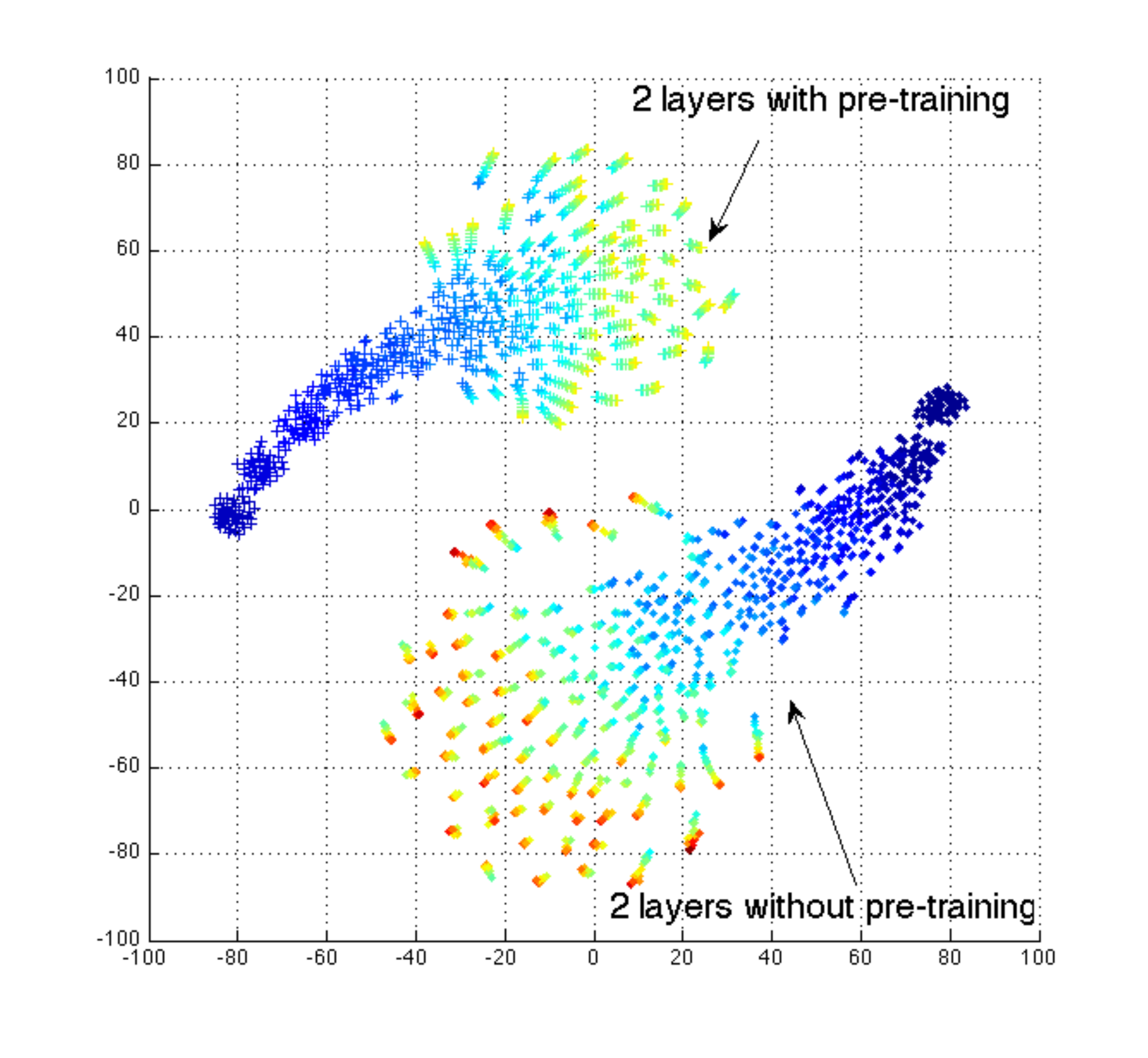}} 

\vspace*{-10mm}
\resizebox{0.6\textwidth}{!}{\includegraphics{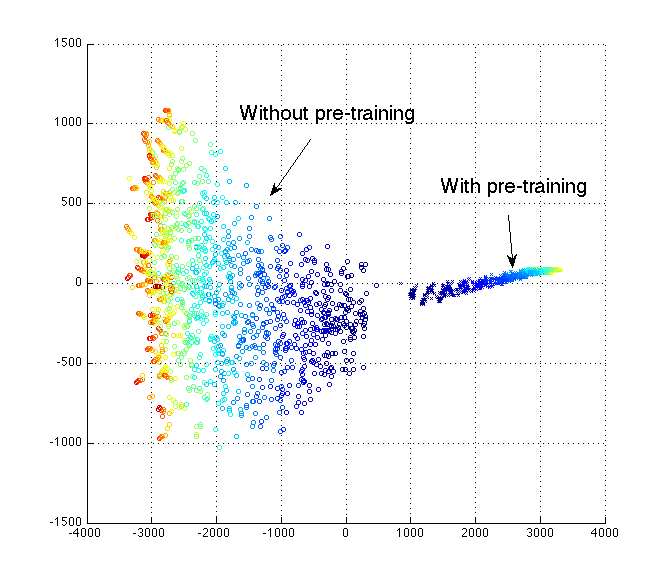}}
\end{center}
\vspace*{-5mm}
\caption{\sl Two-dimensional non-linear projection of the space of functions
visited by artificial neural networks during training. Each cross or diamond
or circle represents a neural network at some stage during its training, with color
indicating its age (number of examples seen), starting from blue and moving
towards red. Networks computing a similar function (with similar response
to similar stimuli) are nearby on the graph. Top figure uses t-SNE for dimensionality
reduction (insists on preserving local geometry) while the bottom figure uses Isomap
(insists on preserving global geometry and volumes). The vertical crosses (top figure)
and circles (bottom figure) are networks
trained from random initialization, while the diamonds (top figure) and
rotated crosses (bottom figure) are networks
with unsupervised pre-training initialization.
\label{fig:DL-localmin}
}
\end{figure}

In~\citet{Erhan+al-2010} we went further in an attempt to understand this
training difficulty and studied the trajectory of deep neural networks
during training, in function space. Such trajectories are illustrated
in Figure~\ref{fig:DL-localmin}. Each point in the trajectory corresponds
to a particular neural network parameter configuration and is visualized
as a two-dimensional point as follows. First, we approximate the function
computed by a neural network non-parametrically, i.e., by the outputs
of the function over a large test set (of 10000 examples). We consider that
two neural networks behave similarly if they provide similar answers on
these test examples. We cannot directly use the network parameters to compare neural
networks because the same function can be represented in many different
ways (e.g., because permutations of the hidden neuron indices would yield
the same network function).
We therefore associate each network with a very long vector containing
in its elements the concatenation of the network outputs on the test examples. This vector
is a point in a very
high-dimensional space, and we compute these points for all the networks
in the experiment. We then learn a mapping from these points to 2-dimensional
approximations, so as to preserve local (and sometimes global) structure as
much as possible, using non-linear dimensionality reduction methods
such as t-SNE~\citep{VanDerMaaten08} or Isomap~\citep{Tenenbaum2000-isomap}.
Figure~\ref{fig:DL-localmin} allows us to draw a number of interesting
conclusions:
\begin{enumerate}
\item {\bf Observation \nO{obs:many-localmin}}. No two trajectories end up in the same local minimum. This suggests that
the number of functional local minima (i.e. corresponding to different
functions, each of which possibly corresponding to many instantiations in
parameter space) must be huge. 
\item {\bf Observation \nO{obs:unreachable}}. A training trick (unsupervised pre-training) which changes
the initial conditions of the descent procedure allows one to reach much better local minima,
and these better local minima do not appear to be reachable by chance alone (note how the regions
in function space associated with the two ``flowers'' have no overlap at all, in fact being
at nearly 90 degrees from each other in the high-dimensional function space).
\end{enumerate}

Starting from the {\bf Local Descent Hypothesis}, 
{\bf Observation O\ref{obs:many-localmin}} and {\bf Observation O\ref{obs:unreachable}} bring us
to the formulation of a new hypothesis regarding not only
artificial neural networks but also humans:\\

\begin{center}
\framebox{
\begin{minipage}{0.8\linewidth}
{\bf Local Minima Hypothesis.} Learning of a single human learner
is limited by effective local minima.
\end{minipage}
}\\
\end{center}

We again used the phrase ``single human learner'' because later in this
paper we will hypothesize that a collection of human learners and the
associated evolution of their culture can help to get out of 
what would otherwise be effective local minima.

Combining the above observations with the worse results sometimes observed when
training deeper architectures ({\bf Observation O\ref{obs:deep-harder}}, discussed above), 
we come to the following hypothesis.\\

\begin{center}
\framebox{
\begin{minipage}{0.8\linewidth}
{\bf Deeper Harder Hypothesis.} 
The detrimental effect of local minima tends to be more pronounced when
training deeper architectures (by an optimization method based on iteratively descending
the training criterion).
\end{minipage}
}\\
\end{center}

Finally, the presumed ability of deeper architectures to represent higher-level
abstractions more easily than shallow ones (see~\citet{Bengio-2009} and discussion 
in Section~\ref{sec:efficiency-repr}) leads us to a human analogue 
of the {\bf Deeper Harder Hypothesis},
which refines the {\bf Local Minima Hypothesis}:\\

\begin{center}
\framebox{
\begin{minipage}{0.8\linewidth}
{\bf Abstractions Harder Hypothesis.}
A single human learner is unlikely to discover high-level
abstractions by chance because these are represented by a deep sub-network
in the brain.
\end{minipage}
}\\
\end{center}

Note that this does not prevent some high-level abstractions to be
represented in a brain due to innate programming captured in the genes,
and again the phrase {\em single human learner} excludes the effects
due to culture and guidance from other humans, which is the subject
of the next section.

\section{Brain to Brain Transfer of Information to Escape Local Minima}
\label{sec:b2b}

If the above hypotheses are true, one should wonder how humans still manage to
learn high-level abstractions. We have seen that much better solutions can
be found by a learner if it is initialized
in an area from which gradient descent leads to a good solution,
and genetic material might provide enough of a good starting point and
architectural constraints to help learning of some abstractions. For example,
this could be a plausible explanation for some visual abstractions (including
simple face detection, which newborns can do) and visual
invariances, which could have had the chance to be discovered by evolution
(since many of our evolutionary ancestors share a similar visual system).
Recent work on learning algorithms for computer vision also suggest that
architectural constraints can greatly help performance of a deep 
neural network~\citep{Jarrett-ICCV2009}, to the point where even
random parameters in the lower layers (along with appropriate connectivity)
suffice to obtain reasonably good performance on simple object recognition
tasks.

\subsection{Labeled Examples as Hints}

However, many of the abstractions that we master today have only recently
(with respect to evolutionary scales) appeared in human cultures, so they
could not have been genetically evolved: each of them must have been discovered by at
least one human at some point in the past and then been propagated or
improved as they were passed from generation to generation. We will return
later to the greater question of the evolution of ideas and abstractions
in cultures, but let us first focus on the mechanics of communicating
good synaptic configurations from one brain to another. Because we have
a huge number of synapses and their values only make sense in the context
of the values of many others, it is difficult to imagine how the recipe for
defining individual abstractions could be communicated from one individual
to another in a direct way (i.e. by exchanging synaptic values). Furthermore, 
we need to ask how the hypothesized mechanism could help to escape effective local minima
faced by a single learner. 

The main insight to answering this question may come from 
{\bf Observation O\ref{obs:layer-hints}}
and {\bf Observation O\ref{obs:sup-easier}}.
Training a single hidden layer neural network (supervised or unsupervised) is much
easier than training a deeper one, so if one can provide a hint as to the
function that deeper layers (corresponding to higher-level abstractions)
should capture, then training would be much easier. In the extreme, specifying
how particular neurons should respond in specific instances is akin to supervised learning.

\begin{figure}[H]
\vspace*{-5mm}
\centerline{\resizebox{0.9\textwidth}{!}{\includegraphics{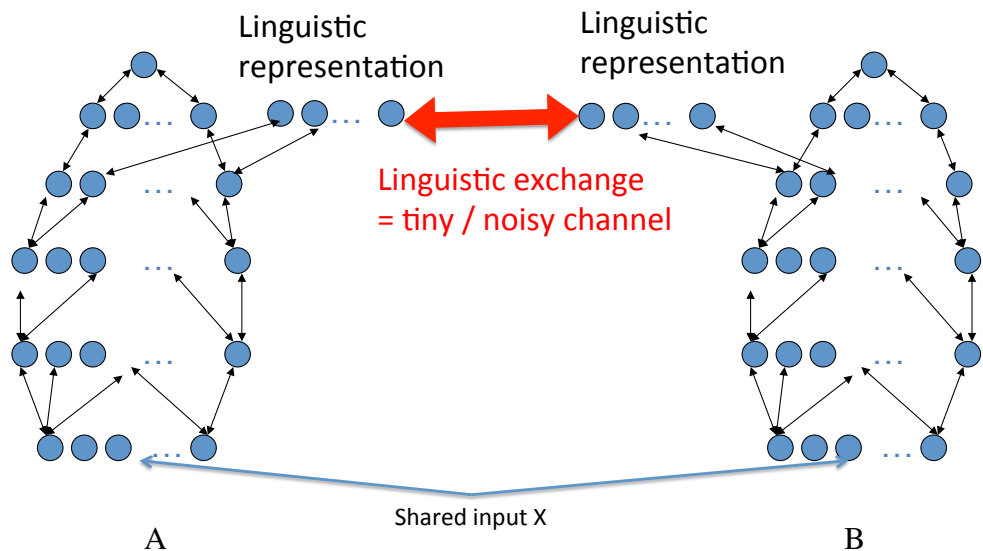}}}
\vspace*{-25mm}
\hspace*{38mm}A\hspace*{90mm}B
\vspace*{2mm}
\caption{\sl Illustration of the communication between brains, typically 
through some language, in a way that can give hints to higher levels of
one brain of how the concepts are represented in higher levels of another brain.
Both learners see shared input $X$, and say that A produces an utterance
(from its language related areas) that is strongly associated with A's
high-level state of mind as a representation of $X$. B also sees this utterance
as an input (that sets B's current linguistic representation units),
that it tries to predict from its internal representations of $X$.
Turns may change with B speaking and A listening, so that both get
a sense of the explanation of $X$ that the other is forming in its
respective state of mind.
\label{fig:communicating-brains}
}
\end{figure}

Based on these premises, the answer that we propose relies on learning agents exchanging bits
of information in the presence of a shared percept.
Communicating about the presence of a concept in a sensory percept is something
that humans do, and benefit from since their youngest age. The situation
is illustrated in Figure~\ref{fig:communicating-brains}.

\subsection{Language for Supervised Training}

A very simple schema that would
help to communicate a concept\footnote{i.e., communicate the characterization
of the concept as a function that associates an indicator of its presence
with all the sensory configurations that are compatible with it.} 
from one brain to another is one
in which there are many encounters between pairs of learners.
In each of them, two learners are faced with a similar percept (e.g.,
they both see the same scene) and they exchange bits of
information about it. These bits can for example be indicators
of the presence of high-level concepts in the scene. These
indicators may reflect the neural activation associated with
these high-level concepts. In humans, these bits of information 
could be encoded through a linguistic convention that helps
the receiver of the message interpret them in terms of 
concepts that it already knows about. One of the most primitive
cases of such a communication scenario could occur with
animal and human non-verbal communication. For example, an adult animal
sees a prey that could be dangerous and emits a danger signal 
(that could be innate) that a young animal could use as a supervised
training signal to associate the prey to danger. 
Imitation is a very common form of learning and teaching, prevalent
among primates, and by which the learner associates contexts
with corresponding appropriate behavior.
A richer
form of communication, which would already be useful, would
require simply {\em naming} objects in a scene. Humans
have an innate understanding of the {\em pointing gesture}
that can help identify which object in the scene is being named.
In this way, the learner could develop a repertoire of object
categories which could become handy (as intermediate concepts)
to form theories about
the world that would help the learner to survive better. Richer
linguistic constructs involve the {\em combination of concepts}
and allow the agents to describe relations between objects, actions and events,
sequences of events (stories), causal links, etc., which are even more useful
to help a learner form a powerful model of the environment.

This brings us to another hypothesis, supported by 
{\bf Observation O\ref{obs:sup-easier}} and {\bf Observation O\ref{obs:layer-hints}}
and following from the
{\bf Abstractions Harder Hypothesis}:\\

\begin{center}
\framebox{
\begin{minipage}{0.8\linewidth}
{\bf Guided Learning Hypothesis.}
A human brain can much more easily learn high-level abstractions
  if guided by the signals produced by other humans, which act as hints
  or indirect supervision for these high-level abstractions.
\end{minipage}
}\\
\end{center}

This hypothesis is related to much previous work in cognitive science, such
as for example {\em cognitive imitation}~\citep{Subiaul-2004}, which has
been observed in monkeys, and where the learner imitates not just a
vocalization or a behavior but something more abstract that corresponds to
a cognitive rule.

\subsection{Learning by Predicting the Linguistic Output of Other Agents}
\label{sec:predict}

How can a human guide another? By encouraging the learner to predict
the ``labels'' that the teacher verbally associates with a given 
input configuration $X$.
In the schema of Figure~\ref{fig:communicating-brains}, it is not necessary
for the emitter (who produces the utterance) to directly provide
supervision to the high-level layers of the receiver (who receives the
communication and can benefit from it). An effect similar to supervised learning
can be achieved {\em indirectly} by simply making sure that the receiver's brain include in
its training criterion the objective of {\em predicting what it observes},
which includes not just $X$ but also the linguistic output of the emitter in the context of the
shared input percept. In fact, with attentional and emotional mechanisms that
increase the importance given to correctly predicting what other humans say
(especially those with whom we have an affective connection), one would
approach even more the classical supervised learning setting.
Since we have already assumed that the training
criterion for human brains involves a term for prediction or maximum
likelihood, this could happen naturally, or be enhanced by innate
reinforcement (e.g. children pay particular attention to the utterances of
their parents). Hence the top-level hidden units $h$ of the receiver would
receive a training signal that would encourage $h$ to become good features
in the sense of being predictive of 
the probability distribution of utterances that are
received (see Figure~\ref{fig:communicating-brains}). This would be naturally
achieved in a model such as the Deep Boltzmann Machine so long as the
higher-level units $h$ have a strong connection to ``language units'' associating
both speech heard (e.g., Wernicke's area) and speech produced (e.g., Broca's area), 
a state of affairs that is consistent
with the global wiring structure of human brains.
The same process could work for verbal or non-verbal
communication, but using different groups of neurons to model the
associated observations. In terms of existing learning algorithms one could
for example imagine the case of a Deep Boltzmann
Machine~\citep{Salakhutdinov+Hinton-2009}: the linguistic units get
'clamped' by the external linguistic signal received by the learner, at the
same time as the lower-level sensory input units get 'clamped' by the
external sensory signal $X$, and that conditions the likelihood gradient
received by the hidden units $h$, encouraging them to model the 
joint distribution of linguistic units and sensory units.

One could imagine many more sophisticated communication schemes that go
beyond the above scenario. For example, there could be a {\em two-way exchange}
of information. It could be that both agents can potentially learn
something from the other in the presence of the shared percept.  Humans
typically possess different views on the world and the two parties in
a communication event could benefit from a two-way exchange. In a sense,
language provides a way for humans to summarize the knowledge collected
by other humans, replacing ``real'' examples by indirect ones, thus
increasing the range of events that a human brain could model.
In that context, it would not be appropriate to simply copy or clone
neural representations from one brain to another, as the learner must
somehow reconcile the indirect examples provided by the teacher with
the world knowledge already represented in the learner's brain.
It could be that there is no pre-assigned role of teacher (as emitter) and 
student (as receiver), but that
depending on the confidence demonstrated by each agent for each particular
percept, one pays more or less attention to the communicated output
of the other. It could be that some aspects of the shared percept are well
mastered by one agent but not the other, and vice-versa. Humans have the
capability to know that some aspect of a situation is surprising (they would
not have predicted it with high probability) and then they should rationally
welcome ``explanations'' provided by others.  A way to make the
diffusion of useful knowledge more efficient is for the communicating
agents to keep track of an estimated degree of ``authority'' or ``credibility''
of other agents. One would imagine that parents and older individuals in a human group 
would by default get more credit, and one of the products of human social systems 
is that different individuals acquire more or less authority and credibility.
For example, scientists strive to maximize their credibility through very
rigorous communication practices and a scientific method that insists on
verifying hypotheses through experiments designed to test them.

\subsection{Language to Evoke Training Examples at Will}

Even more interesting scenarios that derive from linguistic abilities
involve our ability to {\em evoke an input scene}. We do not need to
be in front of danger to teach about it. We can describe a dangerous situation
and mention what is dangerous about it. In this way, the diffusion of
knowledge about the world from human brains to other human brains could
be made even more efficient. 

The fact that verbal and non-verbal communication between animals and humans
happens through a noisy and very bandwidth-limited channel is important to keep in mind.
Because very few bits of information can be exchanged, only
the most useful elements should be communicated. If the objective is only to
maximize collective learning, it seems that there is no point in
communicating something that the receiver already knows. However, there may be other
reasons why we communicate, such as for smoothing social interactions,
acquiring status or trust, coordinating collective efforts, etc.

Note that it is not necessary for the semantics of language to have been
defined a priori for the process described here to work. Since each learning agent
is trying to predict the utterances of others (and thus, producing similar
utterances in the same circumstances), the learning dynamics should converge
towards one or more languages which become attractors for the learning agents:
the most frequent linguistic representation of a given percept $X$ among
the population will tend to gradually dominate in the population. If encounters
are not uniformly random (e.g., because the learning agents are geographically
located and are more likely to encounter spatially near neighbors), then there
could be multiple attractors simultaneously present in the population, i.e.,
corresponding to multiple spatially localized languages.

\begin{figure}[H]
\vspace*{-5mm}
\centerline{\resizebox{0.9\textwidth}{!}{\includegraphics{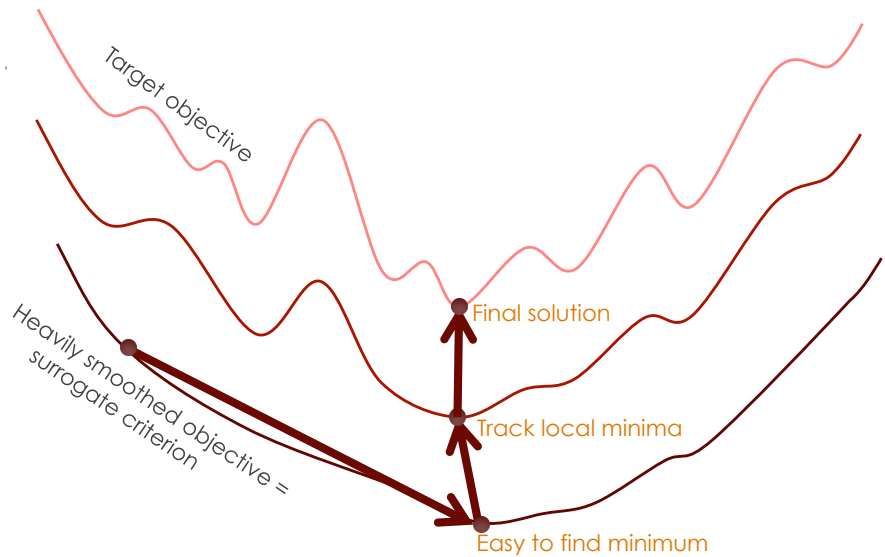}}}
\vspace*{-10mm}
\caption{\sl A general strategy to reduce the impact of local minima
is followed in {\em continuation methods} and {\em simulated annealing}.
The idea is to consider a sequence of optimization problems that start
with an easier one for which it is easy to find a global optimum (not
corresponding to solving the actual problem of interest, though), with the 
sequence of problems ending up in the problem of interest, each time starting at the solution
previously found with an easier problem and tracking local minima
along the way. It was hypothesized~\citep{Bengio-2009} and demonstrated with
artificial neural networks
that following a curriculum could help learners thus find better solutions
to the learning problem of interest.
\label{fig:continuation}
}
\end{figure}

\subsection{Connection with Curriculum Learning}

The idea that learning can be improved by guiding it, by properly choosing
the sequence of examples seen by the learner, was already explored in the 
past. It was first proposed as a practical way to train animals through {\em shaping}
\citep{Skinner1958,Peterson2004}, as a way to ease simulated learning of more
complex tasks~\citep{Elman93,Krueger+Dayan-2009,Bengio+al-2009} by building
on top of easier tasks. An interesting hypothesis introduced in~\citet{Bengio+al-2009}
is that a proper choice of training examples can be used to approximate a complex
training criterion\footnote{the training criterion is here seen as a function of the learned parameters,
as a sum of an error function over a training distribution of examples.}
fraught with local minima with a smoother one (where, e.g., only prototypical
examples need to be shown to illustrate the ``big picture''). Gradually
introducing more subtle examples and building on top of the already understood
concepts is typically done in pedagogy. \citet{Bengio+al-2009} propose
that the learner goes through a sequence of gradually more difficult learning
tasks, in a way that corresponds in the optimization literature to
a {\em continuation method} or an {\em annealing method}, allowing one to 
approximately discover global minima (or much better local minima),
as illustrated in Figure~\ref{fig:continuation}.
Interestingly, it was recently observed experimentally that humans use
a form of curriculum learning strategy (starting from easier examples and
building up) when they are asked to teach a concept to a robot~\citep{Khan+Zhu+Mutlu-2011}.
\citet{Khan+Zhu+Mutlu-2011} also propose a statistical explanation why
a curriculum learning strategy can be more successful, based on the
uncertainty that the learner has about the relevant factors explaining the 
variations seen in the data. If these theories are correct, an individual
learner can be helped (to escape local minima or converge faster to better
solutions) not only by showing examples of abstractions not yet mastered
by the learner, but also by showing these well-chosen examples in an appropriate sequence.
This sequence corresponds to a curriculum that helps the learner build higher-level abstractions
on top of lower-level ones, thus again defeating some of the difficulty
believed to exist in training a learner to capture higher-level abstractions.

\section{Memes, Cross-Over, and Cultural Evolution}

In the previous section we have proposed a general mechanism by which
knowledge can be {\em transmitted} between brains, without having to
actually {\em copy synaptic strengths}, instead taking advantage of the
learning abilities of brains to transfer concepts via examples. We hypothesized that such
mechanisms could help an {\em individual learner} escape an effective local
minimum and thus construct a better model of reality, when the learner is
guided by the hints provided by other agents about relevant abstractions.
But the knowledge had to come from another agent. Where did this knowledge
arise in the first place? This is what we discuss here.

\subsection{Memes and Evolution from Noisy Copies}

Let us first step back and ask how ``better''\footnote{``better'' in the sense of the
survival value they provide, and how well they allow their owner to understand
the world around them. Note how this depends on the context (ecological and
social niche) and that there may be many good solutions.} brains could arise. The most plausible explanation is that
better brains arise due to some form of search or optimization
(as stated in the {\bf Optimization Hypothesis}), in the huge
space of brain configurations (architecture, function, synaptic strengths).
Genetic evolution is a form of parallel search (with each individual's genome
representing a candidate solution) that occurs on a rather slow time-scale.
Cultural evolution in humans is also a form of search, in the space
of ideas or {\em memes}~\citep{Dawkins-1976}. A meme is a unit of selection
for cultural evolution. It is something that can be copied from one mind to
another. Like for genes, the copy can be imperfect. 
Memes are analogous to genes in the context of cultural evolution~\citep{Distin-2005}.
Genes and memes have co-evolved, although it appears that cultural evolution
occurs on a much faster scale than genetic evolution. Culture allows
brains to modify their basic program and we propose that culture
also allows brains to go beyond what a single individual can achieve by 
simply observing nature. Culture allows
brains to take advantage of knowledge acquired by other brains elsewhere
and in previous generations. 

\begin{figure}[H]
\vspace*{-5mm}
\centerline{\resizebox{0.75\textwidth}{!}{\includegraphics{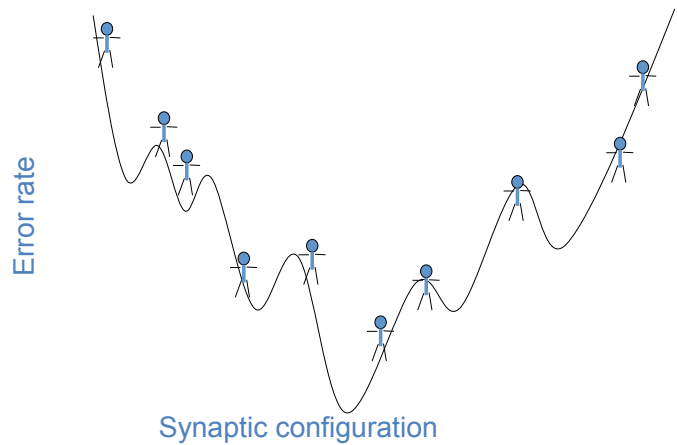}}}
\vspace*{-10mm}
\caption{\sl Illustration of parallel search in the space of synaptic
configurations by a population of learners. Some learners start
from configurations which happen to lead to a better solution
when descending the training criterion.
\label{fig:parallel-search}
}
\end{figure}

To put it all together, the knowledge acquired by an individual brain
combines four levels of adaptation: {\bf genetic evolution} (over hundreds of
thousands of years or more), {\bf cultural evolution} (over dozens, hundreds or
thousands of years), {\bf individual learning} and discovery (over minutes,
hours and days) and {\bf inference} (fitting the state of mind to the observed perception,
over split seconds or seconds).  In all four cases, a form of adaptation is at play,
which we hypothesize to be associated with a form of approximate
optimization, in the same sense as stated in the {\bf Optimization Hypothesis}.
One can also consider the union of all four adaptation processes
as a global form of evolution and adaptation (see the work
of~\citet{Hinton+Nowlan-89} on how learning can guide evolution
in the style of Baldwinian evolution). Whereas genetic
evolution is a form of parallel search (many individuals carry different
combinations and variants of genes which are evaluated in parallel)
and we have hypothesized that individual learning is a local search
performing an approximate descent ({\bf Local Descent Hypothesis}), what
about cultural evolution? Cultural evolution is based on individual
learning, on learners trying to predict the behavior and speech output
of individuals, as stated in the {\bf Guided Learning Hypothesis}. Even though
individual learning relies on a local descent to gradually improve a single brain, 
when considering the graph of interactions between humans
in an evolving population, one must conclude that cultural evolution,
like genetic evolution, is a form of parallel search, as illustrated
in Figure~\ref{fig:parallel-search}.

The most basic working principle of evolution is the {\em noisy copy}
and it is also at work in cultural evolution: a meme can be noisily copied from
one brain to another, and the meme can sometimes be slightly modified
in the process\footnote{Remember that a meme is copied in a process
of teaching by example which is highly stochastic, due to the randomness
in encounters (in which particular percepts serve as examples of the meme)
and due to the small number of examples of the meme. This creates a highly
variable randomly distorted version of the meme in the learner's brain.}. 
A meme exists in a human's brain as an aspect of the dynamics of the brain's neural
network, typically allowing the association of words in language (which are encoded
in specific areas of the brain) with high-level abstractions learned
by the brain (which may be encoded in other cortical areas, depending the
semantics of the meme). The meme is activated when neural configurations
associated with it arise, and different memes are also connected to each
other in the sense of having a high probability of being associated together and
echoing each other through thoughts, reasoning, or planning.

Selective pressure then does the work of exponentially
increasing the presence of successful memes in the population, by
increasing the chances that a successful meme be copied in comparison
with a competing less successful meme. This may happen simply because
a useful meme allows its bearer to survive longer, communicate with
more individuals, or because better ideas are promoted~\footnote{Selfish
memes~\citep{Dawkins-1976,Distin-2005} may also strive in a population: they do not
really help the population but they nonetheless maintain themselves in
it by some form of self-promotion or exploiting human weaknesses.}.
With genetic evolution, it is necessary to copy a whole genome
when the individual bearing it is successful. Instead, cultural evolution
in humans has mechanisms to evaluate an {\em individual meme} and selectively promote
it. Good ideas are more likely to be the subject of discussion in
public communication, e.g., in the public media, or even better
in scientific publications. Science involves powerful mechanisms
to separate the worth of a scientist from the worth of his or her ideas
(e.g. through independent replication of experimental results
or theoretical proofs, or through blind reviewing). That may
explain why the pace of evolution of ideas has rapidly increased since
the mechanisms for scientific discovery and scientific dissemination
of memes have been put in place. The fact that a good idea can stand on its own
and be selected for its own value means that the {\em selective pressure}
is much more efficient because it is less hampered by the noisy evaluation
that results when fitness is assigned to a whole individual, that integrates
many memes and genes. 

In this context the premium assigned to novelty in some cultures,
in particular in scientific research, makes sense as it favors novel memes
that are farther away from existing ones. By increasing the degree of
exploration through this mechanism, one might expect that it would yield
more diversity in the solutions explored, and thus more efficient
search (finding good ideas faster) may be achieved with appropriate
amounts of this premium for novelty.

\subsection{Fast-Forward with Divide-and-Conquer From Recombination}

But if evolution only relied on the noisy copy principle, then it could
only speed-up search at best linearly with respect to the number of individuals
in a population. Instead of trying $N$ random configurations with $N$ individuals
and picking the best by selective pressure, a population with $M>N$ individuals
would discover a good selection $M/N$ times faster in average. 
This is useful but we hypothesize that it would not be enough to make a real dent
in the optimization difficulty due to a huge number of poor local minima
in the space of synaptic configurations. In fact,
evolution has discovered an evolutionary mechanism which can yield much
larger speed-ups, and is based on {\em sexual reproduction} in the case
of genetic evolution. With sexual reproduction, we have an interaction
between two parent individuals (and their associated candidate configurations),
and we mix some of the genes of one with some of the genes of the other
in order to create new combinations that are not near-neighbors of either
parent. This is very different from a simple parallel search because
it can explore new configurations beyond local variations around the
randomly initialized starting stock. Most importantly, a {\em recombination operator} can
{\em combine good, previously found, sub-solutions}. Maybe your father
had exceptionally good genes for eyes and your mother 
exceptionally good genes for ears, and with about 25\% probability
you could get both, and this may confer you with an advantage that
no one had had before. This kind of transformation of the population
of candidate configurations is called a {\em cross-over operator}
in the genetic algorithms literature~\citep{Holland75}. Cross-over is
a recombination operator: it can create new candidate
solutions by combining {\em parts} of previous candidate solutions.
Cross-over and other operators that {\em combine existing parts of solutions to form
new candidate solutions} 
have the potential for a much greater speed-up than simple
parallelized search based only on individual local descent (noisy copy).
This is because such operators can potentially
exploit a form of {\em divide-and-conquer}, which, if well done,
could yield exponential speed-up. For the divide-and-conquer aspect
of the recombination strategy to work, it is best if {\em sub-solutions
that can contribute as good parts to good solutions receive a high
fitness score}. As is well known in computer science, divide-and-conquer
strategies have the potential to achieve an {\em exponential speedup}
compared to strategies that require blindly searching through potential
candidate solutions (synaptic configurations, here). The exponential
speedup would be achieved if the optimization of each of the combined
parts (memes) can be done independently of the others. In practice,
this is not going to be the case, because memes, like genes, only
take value in the context and presence of other memes in the individual
and the population.

The success rate of recombination is also important
i.e., what fraction of the recombination offsprings are viable?
The {\em encoding} of information into genes has a great influence
on this success rate as well as on the fitness assigned to
good sub-solutions. We hypothesize that memes are particularly
good units of selection in these two respects: they are by definition
the units of cultural information that can be meaningfully 
recombined to form new knowledge. All these ideas are
summarized in the following hypothesis.\\

\begin{center}
\framebox{
\begin{minipage}{0.8\linewidth}
{\bf Memes Divide-and-Conquer Hypothesis.} 
Language, individual learning, and the recombination
  of memes constitute an efficient evolutionary
  recombination operator, and this  
  gives rise to rapid search in the space of memes,
  that helps humans build up better 
  high-level internal representations of their world.
\end{minipage}
}\\
\end{center}

\subsection{Where do New Ideas Come from?}

Where do completely new ideas (and memes) emerge? According to the views
stated here, they emerge from two intertwined effects. On the one hand, our
brain can easily combine into new memes different memes which it inherited
from other humans, typically through linguistic communication and
imitation. On the other hand, such recombination as well as other creations
of new memes must arise from the optimization process taking place in a
single learning brain, which tries to reconcile all the sources of evidence that it
received into some kind of unifying theory. This search is local in
parameter space (synaptic weights) but can involve a stochastic search in
the space of neuronal firing patterns (state of mind). For example, in a Boltzmann machine,
neurons fire randomly but with a probability that depends on the
activations of other connected neurons, and so as to explore and reach more
plausible ``interpretations'' of the current and past observations (or ``planning''
for future actions in search for a sequence of decisions that would give
rise to most beneficial outcomes), given
the current synaptic strenghts. In this stochastic exploration, new
configurations of neuronal activation can randomly arise and if these do a
better job of explaining the data (the observations made), then synaptic
strengths will change slightly to make these configurations more likely in
the future.  This is already how some artificial neural networks learn and
``discover'' concepts that explain their input. In this way, we can see
``concepts'' of edges, parts of face, and faces emerge from a deep Bolzmann
machine that ``sees'' images of faces~\citep{HonglakL2009}.

What this means is that the recombination operator for memes is doing
much more than recombination in the sense of cutting and pasting
parts together. It does that but it is also possible for the new combinations
to be {\em optimized} in individual brains (or even better, by groups
who create together) so as to better fit the empirical 
evidence that each learner has access to. This is related to the
ideas in~\citet{Hinton+Nowlan-89} where a global search (in their
case evolution) is combined with a local search (individual learning).
This has the effect of smoothing the fitness function seen by the
global search, by allowing half-baked ideas (which would not work
by themselves) to be tuned into working ones, i.e., replacing
the needle in the haystack by a glowing needle in the haystack,
which is much easier to find.



\section{Conclusion and Future Work}

To summarize, motivated by theoretical and empirical work on
Deep Learning, we developed a theory starting from the hypothesis
that high-level abstractions are difficult to
learn because they need to be represented with highly non-linear computation
associated with enough levels of representation, and that this difficulty
corresponds to the learner getting stuck around effective local minima.
We proposed and argued that other learning agents can provide new examples to the learner
that effectively change the learner's training criterion into one where these
difficult local minima are not minima anymore. This happens because the
communications from other agents can provide a kind of indirect supervision
to higher levels of the brain, which makes the task of discovering 
explanatory factors of variation (i.e., modeling the rest of the observed
data) much easier. Furthermore, this brain-to-brain communication
mechanism allows brains to recombine nuggets of knowledge called memes.
Individual learning corresponds to searching for such recombinations and other
variations of memes that are good at explaining the data observed
by learners. In this way, new memes are created that can be disseminated
in the population if other learning agents value them, creating
cultural evolution. Like genetic evolution, cultural evolution efficiently
searches (in the space of memes, rather than genes) thanks to parallelism,
noisy copying of memes, and creative recombination of memes. We hypothesize
that this phenomenon provides a divide-and-conquer advantage that yields 
much greater speedup in the optimization performed, compared to the linear speedup
obtained simply from parallelization of the search across a population.

A lot more needs to be done to connect the above hypotheses with the wealth
of data and ideas arising in the biological social sciences. They can
certainly be refined and expanded into more precise statements. Of central
importance to future work following up on this paper is how one could go
and {\em test these hypotheses}.  Although many of these hypotheses agree
with common sense, it would be worthwhile verifying them empirically,
to the extent this is possible.  It is also quite plausible that many
supporting experimental results from neuroscience, cognitive science,
anthropology or primatology already exist that support these hypotheses,
and future work should cleanly make the appropriate links.

To test the {\bf Optimization Hypothesis} would seem to require estimating
a criterion (not an obvious task) and verifying that learning improves it in average. 
A proxy for this criterion (or its relative change, which is all we care about, here)
might be measurable in the brain itself, for example
by measuring the variation in the presence of reward-related molecules or the activity
of neurons associated with reward. The effect of learning could be tested
with a varying number of training trials with respect to a rewarding task.

If the {\bf Optimization Hypothesis} is considered true, testing the additional
assumptions of the {\bf Local Descent Hypothesis} is less obvious because it is difficult
to measure the change in synaptic strengths in many places. However, a form of {\em stability}
of synaptic strengths is a sufficient condition to guarantee that the optimization has
to proceed by small changes.

There is already evidence for the {\bf Deep Abstraction Hypothesis} in the
visual and auditory cortex, in the sense that neurons that belong to areas
further away from the sensory neurons seem to perform a higher-level
function. Another type of evidence comes from the time required to solve
different cognitive tasks, since the hypothesis would predict that tasks
requiring computation for the detection of more abstract concepts would
require longer paths or more ``iterations'' in the recurrent neural network
of the brain.

The {\bf Local Minima Hypothesis} and the {\bf Abstractions Harder Hypothesis}
are ethically difficult to test directly but
are almost corollaries of the previous hypotheses. An indirect source of
evidence may come from raising a primate without any contact with other
primates nor any form of guidance from humans, and measure the effect on
operational intelligence at different ages. One problem with such an
experiment would be that other factors might also explain a poor
performance (such as the effect of psychological deprivation from social
support, which could lead to depression and other strong causes of poor
decisions), so the experiment would require a human that provides warmth
and caring, but no guidance whatsoever, even indirectly through
imitation. Choosing a more solitary species such as the orangutan would
make more sense here (to reduce the impacts due to lack of social
support). The question is whether the tested primate could learn to survive
as well in the wild as other primates of the same species. 

The {\bf Guided Learning Hypothesis} could already be supported by empirical
evidence of the effect of education on intelligence, and possibly by
observations of feral (wild) children. The important point
here is that the intelligence tests chosen should not be about reproducing
the academic knowledge acquired during education, but about decisions where
having integrated knowledge of some learned high-level abstractions could
be useful to properly interpret a situation and take correspondingly
appropriate decisions. Using computational simulations with artificial
neural networks and machine learning one should also test the validity
of mechanisms for ``escaping'' local minima thanks to ``hints'' from
another agent.

The {\bf Memes Divide-and-Conquer Hypothesis} could probably be best tested by
computational models where we simulate learning of a population of agents
that can share their discoveries (what they learn from data) by
communicating the high-level abstractions corresponding to what they
observe (as in the scenario of Section~\ref{sec:b2b} and
Figure~\ref{fig:communicating-brains}).  The question is whether one could
set up a linguistic communication mechanism that would help this population
of learners converge faster to good solutions, compared to a group of
isolated learning individuals (where we just evaluate a group's
intelligence by the fitness, i.e. generalization performance, of the
best-performing individual after training). Previous computational
work on the evolution of language is also relevant, of course.
If such algorithms would work,
then they could also be useful to advance research in machine learning and
artificial intelligence, and take advantage of the kind of massive and
loose parallelism that is more and more available (to compensate for a
decline in the rate of progress of the computing power accessible by a
single computer core). This type of work is related to other research on
algorithms inspired by the evolution of ideas and culture (see the {\em
  Wikipedia} entry on {\em Memetic Algorithms} and~\citep{Moritz-1990,Hutchins+Hazlehurst-95,Hutchins+Hazlehurst-02}).

If many of these hypotheses (and in particular this last one) are true, then
we should also draw conclusions regarding the {\em efficiency of cultural
  evolution} and how different {\em social structures} may influence that
efficiency, i.e., yield greater group intelligence in the long run. Two
main factors would seem to influence this efficiency: (1), the efficiency
of exploration of new memes in the society, and (2), the rate of spread of
good memes. Efficiency of exploration in meme-space would be boosted by a
greater investment in scientific research, especially in high-risk high
potential impact areas. It would also be boosted by encouraging diversity
it all its forms because it would mean that individual humans explore a
less charted region of meme-space.  For example, diversity would be boosted
by a non-homogeneous education system, a general bias favoring openness
to new ideas and multiple schools of thought (even if they disagree),
and more generally to marginal beliefs and individual differences.
The second factor, the rate of spread of good memes, would be boosted
by communication tools such as the Internet, and in particular by open
and free access to education, information in general, and
scientific results in particular. The investment in education would
probably be one of the strongest contributors of this factor,
but other interesting contributors would be social structures making it easy for
every individual to disseminate useful memes, e.g., to publish
on the web, and the operation of non-centralised systems of rating
what is published (whether this is scientific output or individual
blogs and posts on the Internet), 
helping the most interesting new ideas to bubble up and spread faster,
and contributing both to diversity of new memes and more efficient dissemination
of useful memes. Good rating systems could help humans to detect
selfish memes that ``look good'' or self-propagate easily
for the wrong reasons (like cigarettes
or sweets that may be detrimental to your health even though many people
are attracted to them), and the attempts at objectivity and replicability
that scientists are using may help there. 

\subsubsection*{Acknowledgements}

The author would like to thank Caglar Gulcehre, Aaron Courville, Myriam Côté, 
and Olivier Delalleau for useful feedback, as well as NSERC and the Canada Research
Chairs for funding.

\bibliographystyle{natbib}
\bibliography{strings,ml,aigaion}

\end{document}